\documentclass{article}
\usepackage{spconf,amsmath,graphicx}
\usepackage{color}
\usepackage{times}
\usepackage{epsfig}
\usepackage{epstopdf}
\usepackage{graphicx}
\usepackage{amsmath}
\usepackage{amssymb}
\usepackage{bm}
\usepackage{subfigure}
\usepackage{multirow}
\usepackage{diagbox}
\usepackage{multicol}
\usepackage{breqn}
\usepackage{algorithm}
\usepackage{algpseudocode}

\title{Calibration of depth cameras using denoised depth images}
\def\hspa{\hspace{0.5in}}
\def\hspb{\hspace{0.7in}}
\name{Ramanpreet Singh Pahwa$^1$ \hspa Minh N. Do$^1$ \hspa Tian Tsong Ng$^2$ \hspa Binh-Son Hua$^3$}
\address{$^1$University of Illinois, Urbana-Champaign, USA \hspb
		 $^2$Institute for Infocomm Research, Singapore \\
		 $^3$National University of Singapore}

\begin{document}
\ninept

\maketitle
\begin{abstract}
Depth sensing devices have created various new applications in scientific and commercial research with the advent of Microsoft Kinect and PMD (Photon Mixing Device) cameras. Most of these applications require the depth cameras to be pre-calibrated. However, traditional calibration methods using a checkerboard do not work very well for depth cameras due to the low image resolution. In this paper, we propose a depth calibration scheme which excels in estimating camera calibration parameters when only a handful of corners and calibration images are available. We exploit the noise properties of PMD devices to denoise depth measurements and perform camera calibration using the denoised depth as additional set of measurements. Our synthetic and real experiments show that our depth denoising and depth based calibration scheme provides significantly better results than traditional calibration methods.
\end{abstract}

\begin{section}{Introduction}
An important recent development in visual information acquisition is the emerging low-cost and fast cameras for measuring depth. With the advent of Microsoft Kinect  \cite{kinect_website}, PMD CamCube $3.0$ \cite{pmd_website} and WAVI Xtion \cite{wavi_xtion_website}, the depth cameras are being used extensively in applications such as gaming and virtual reality. While PMD camera measures the time of flight (TOF) of infrared light, Kinect uses structured light to estimate depth at each pixel. With the development of these depth cameras, the structural information about the scene can be captured at high speed, and it can be incorporated in many applications due to their portibility. Obtaining such information is crucial in many 3D applications; examples include image based rendering \cite{Kolb2009}, 3D reconstruction \cite{Izadi2011}, and motion capture \cite{shotton2013real}.

In order to perform such tasks, depth cameras need to be properly calibrated. Camera calibration refers to performing a set of controlled experiments to determine initial parameters of the camera that affect the imaging process of the scene.  Thus, camera calibration is an extremely important step in 2D and 3D computer vision. Unfortunately, the imaging capabilities of some TOF cameras are very limited when compared to conventional color sensors. They can only provide a low-resolution intensity image and depth map containing significant depth noise. This causes the traditional calibration scheme to be inaccurate. Hence, both the camera calibration and the depth denoising need to be significantly improved to obtain satisfactory calibration results.

In this paper, we propose a novel algorithm that takes in few calibration images and utilizes them to simultaneously denoise and calibrate TOF depth cameras. Our formulation is based on two key elements. First, we use depth planarization in $3$D to denoise the depth at each corner pixel. Then, in the second stage, we use these improved depth measurements along with the corner pixel information to estimate the calibration parameters using a non-linear estimation algorithm. We demonstrate that our framework estimates the intrinsic and extrinsic calibration parameters more accurately using less number of images and corners that are needed for traditional camera calibration. We evaluate our approach on both synthetic dataset where groundtruth information is available, and real data taken from a PMD camera. In both cases, we demonstrate that our proposed framework outperforms traditional calibration technique without significant increase in computational complexity. Moreover, our framework requires less number of images and corners which makes it easier to use for general public.
\end{section}
\begin{section}{Related Work}
\noindent \textbf{Color camera calibration:} A lot of work has been done in computer vision and photogrammetry community \cite{zhang2000flexible, Maybank1992} to perform color camera calibration. The traditional approaches use a set of checkerboard images taken at various positions and exploit planar geometry to estimate the calibration parameters.

\vspace{1mm}
\noindent \textbf{PMD camera calibration: } Since PMD cameras are relatively new, most of the current approaches borrow heavily from traditional camera calibration technique. Kahlmann \textit{et al.} \cite{kahlmann2006calibration} explore the depth related errors at various exposure times. They use a look-up table to correct for the depth noise. This approach is time consuming and entails creating a look-up table each time. Linder \textit{et al.} \cite{lindner2006lateral} use a controlled set of measurements to perform depth camera calibration. The checkerboard is put on a very precise optical measurement rack which is moved $10cm$ away from camera iteratively and this prior knowledge is used to correct the depth at corner points. Fuchs and Hirzinger \cite{fuchs2008extrinsic} use a color and a depth camera rigidly set up on a robotic arm and move the arm with a pre-determined set of poses to estimate the calibration parameters using a checkerboard. They do not estimate the lens distortion parameters assuming the camera contains insignificant radial and tangential distortion. Beder and Koch \cite{beder2008calibration} estimate the focal length and extrinsic parameters of the  PMD camera using the intensity map and depth measurements from a single checkerboard image. They assume the camera to be distortion free with optical center lying at the image center. 

\vspace{1mm}
\noindent \textbf{Kinect camera calibration: } Kim \textit{et al.} \cite{kim2011depth} present a method to calibrate and enhance depth measurements for Kinect. They project the depth onto color sensor's camera plane and use a Weighted Joint Bilateral Filter considering the color and depth information at the same time to reduce the depth noise. Herrera \textit{et al.} \cite{herrera2011accurate} use a depth and color camera pair to perform camera calibration using a planar checkerboard by utilizing the camera's depth to improve the calibration. However, they assume the depth camera to be distortion free and only estimate two disparity mapping related parameters for the Kinect camera. Hence, their method is unable to estimate the actual intrinsic parameters of the depth camera. In a recent work, Herrera  \textit{et al.}  \cite{herrera2012joint} propose an algorithm that performs calibration with Kinect depth sensor and two color cameras using $60$ checkerboard images. While their algorithm accounts for depth noise, they assume the depth sensor to be distortion free. Our approach closely resembles their approach. However, we use a PMD camera that contains significant photon noise and has a much lower resolution than Kinect. 

Most of these techniques either require multiple cameras or a controlled set-up to exploit some prior knowledge to estimate the calibration parameters. Moreover, most of these approaches ignore lens distortion which is significant in PMD cameras. We aim to provide a simple approach that estimates lens distortion and performs calibration while simultaneously denoising the depth map by exploiting scene planarity using as few images and corners as possible.

\end{section}

\begin{section}{Standard Camera Calibration}\label{Sec_std_calib} 

In this section, we describe the basics of traditional color camera calibration and a commonly used algorithmic approach to estimate the camera calibration parameters.

\vspace{1mm}
\noindent \textbf{Color Camera Calibration Parameters:} The intrinsic calibration matrix of a camera, $\bm{K}$, contains five parameters - focal length in $x$ and $y$ directions, $[f_x,f_y]^\top$; skew $s$; and the location of optical center, $[c_x,c_y]^\top$ as defined in \cite{zhang2000flexible}. The skew is commonly set to zero for non fish-eye lenses. Usually a lens is more ``spherical'' than being perfectly parabolic. This leads to radial distortion. Another common distortion seen in some cameras happens when the sensor and lens do not align properly. This results in tangential distortion. This usually happens due to manufacturing defects where the imaging plane of the camera is not perfectly parallel to the lens. The radial and tangential distortion are normally bundled together as ${\bm{k}_c = [k_1,k_2,k_3,k_4]^\top}$. We represent a 3D point in camera coordinate frame as $\bm{x}_c$. The 3D points are projected onto camera plane at the normalized pixel position, $\bm{x}_n=[x_{n}, y_{n}]^\top$ as:
\begin{equation}\label{2d_projection}
\begin{bmatrix} x_n  \\[0.3em]  y_n  \end{bmatrix} = \begin{bmatrix} x_{c,1}/x_{c,3} 	  \\[0.3em]   x_{c,2}/x_{c,3} \end{bmatrix} 
\end{equation}
The distorted pixel value of this point, $\bm{x}_d$, is obtained after adding the forward distortion model as:
\begin{equation}
\begin{bmatrix} x_d  \\[0.3em]  y_d  \end{bmatrix} =  \begin{bmatrix} x_n (1+k_1r^2+k_2r^4 + 2k_3y_n) + k_4(r^2+2x_n^2)  \\[0.3em] y_n (1+k_1r^2+k_2r^4 + 2k_4x_n) + k_3(r^2+2y_n^2)\end{bmatrix} \\
\end{equation}
Here, $r$ refers to the magnitude of the normalized pixel position. Lets call this function $\it{h}$, i.e., $\bm{x}_d = h(\bm{x}_n,\bm{k}_c)$. Eventually, the final pixel position, $\bm{x}_p$, recorded by the camera is obtained by using the intrinsic calibration matrix as: 
\begin{equation}\label{instrinsic_scaling}
\begin{bmatrix} x_p  \\[0.3em]  y_p \\[0.3em] 1 \end{bmatrix} = \bm{K} \begin{bmatrix}    x_d  \\[0.3em]  y_d \\[0.3em] 1  \end{bmatrix}  ; \qquad\qquad   \bm{K} = \begin{bmatrix}  f_x 	& 	0 	& c_x           \\[0.3em]  0    	&	f_y  	& c_y \\[0.3em]  0  		& 	0 	& 1 \end{bmatrix}
\end{equation}

\vspace{1mm}
\noindent \textbf{Color Camera Calibration Scheme}
There are various ways to perform color camera calibration with lens distortion taken into account. A widely used calibration toolbox \cite{bouguet2004camera} uses a planar checkerboard pattern with $M$ corners to perform the calibration. The user holds a checkerboard in front of the camera and takes $N$ images with the checkerboard held in various positions. The 3D points that lie on the checkerboard are expressed in terms of a world coordinate frame, $\bm{x}_w$. For every image, the two coordinate frames are related via a rotation matrix, $\bm{R}$, and translation vector, $\bm{t}$. 
\begin{equation}\label{eq_world_coord}
\bm{x}_c = \bm{R}\bm{x}_w+\bm{t} 
\end{equation}
Both the rotation matrix and translation vector contain three parameters each. The rotation matrix and translation vector, $\{\bm{R}^j$,$\bm{t}^j\}$, are bundled together for each image and calibrated together with the intrinsic parameters. We denote all the calibration parameters $(\bm{K}, {\bm{k}_c}, \{ \bm{R}^1$, $\bm{t}^1, \bm{R}^2$, $\bm{t}^2, \hdots \bm{R}^N$, $\bm{t}^N \})$ as $\bm{V}$.

\vspace{1mm}
\noindent \textbf{Global Optimization:} The following objective function is used in traditional calibration to obtain the calibration parameters by minimizing the projected 2D distance between the measured corners and projected corners:
\begin{equation}\label{std_calib_eq}
\bm{\hat{V}}= \operatornamewithlimits{argmin}\limits_{\bm{V}} \sum_{j=1}^N\sum_{i=1}^M{\left( ||\bm{x}^{i,j}_p - \bm{x}^{i,j}_{m}||_2^2 )\right)}
\end{equation}
Here, $\bm{x}^{i,j}_p$ refers to the $i^{th}$ corner of $j^{th}$ image projected on the camera plane and $\bm{x}^{i,j}_m$ refers to the actual corresponding measured corner using corner detection algorithm. This is usually solved using a non-linear estimator such as gradient-descent or Levenberg-Marquardt algorithm (LMA) with a user defined Jacobian matrix. 

\vspace{1mm}
\noindent \textbf{Initialization:} Most non-linear solvers such as LMA require a good initialization. The distortion, $\bm{k}_c$, is initialized as zero. A planar homography, per image, between the interior corners of the checkerboard in world coordinate frame and imaging plane is estimated. These matrices are combined together to initialize $\bm{K}$ using Direct Linear Transformation (DLT) algorithm. Then, $\bm{K}$ is used to reinitialize rotation and translation per image individually by decomposing the homography matrices \cite{Hartley2004,Bradski2008}. The extrinsic parameters are usually re-estimated per image individually using LMA for better accuracy. This is known as local optimization.  After performing local optimization, the parameters are bundled together and global optimization is performed on the entire dataset as seen in Eq. (\ref{std_calib_eq}). 


\end{section}

\section{Depth Camera Calibration}
PMD depth cameras not only provide us an estimated intensity image but also another measurable quantity - depth at each pixel. This is the 3D scalar distance between the camera center and the point in 3D corresponding to that pixel. Using Eq. (\ref{eq_world_coord}), we can represent depth as:
\begin{equation}
d = \|\bm{x}_c\|_2  = \sqrt{\|\bm{x}_w\|_2^2 + \|\bm{t}\|_2^2 + 2\bm{t}{^\top}\bm{R}\bm{x}_w}
\end{equation}
We use this additional set of measurements per corner pixel to perform the global optimization  process by minimizing the following function using LMA with a user defined Jacobian matrix.
\begin{equation}\label{our_opti_eq}
\begin{split}
\bm{\hat{V}}= \operatornamewithlimits{argmin}\limits_{\bm{V}}  \sum_{j=1}^N \sum_{i=1}^M  \left( \frac{\|\bm{x}^{i,j}_p - \bm{x}^{i,j}_m\|_2^2}{(\sigma_x^j)^2} + \frac{(d^{i,j}_p - d^{i,j}_m)^2}{(\sigma_d^j)^2}\right)
\end{split}
\end{equation}
Here, $d^{i,j}_p$ refers to the estimated depth of $i^{th}$ corner of $j^{th}$ image and $d^{i,j}_m$ refers to the measured depth by the depth camera. We normalize the error terms in Eq. (\ref{our_opti_eq}) with their respective variances, ($\{(\sigma_x^j)^2, (\sigma^j_d)^2\}$) for every image, as they have different measurement units. 

\vspace{1mm}
\noindent \textbf{Depth noise: }Like every sensing device, PMD also exhibits various error sources which effect the accuracy of depth information captured by it. There are three major sources of error in PMD cameras. First, the wiggling error is caused due the hardware and manufacturing limitations. The outgoing signal is assumed to be perfectly sinosoidal. However, in reality, this signal is more ``box-shaped'' than sinosoidal \cite{lindner2010time}. Second, the flying-pixel error occurs at depth discontinuities. The depth at each pixel is computed by using four readings at each pixel. The information captured at each smart-pixel in PMD can come from either the background and foreground object which leads to an unreliable depth measurement at these pixels. Third, the Poisson-shot noise error occurs due to reflectivity of the scene \cite{lindner2010time}. This inherent noise present in the capturing process leads to an unsteady 3D point cloud. The noise can be partly reduced by spatial averaging using bilateral filters, but we cannot use this process for applications requiring accurate depth map as smoothing a depth map is highly undesirable. Thus, before we use the depth measurements, we pre-process the depth image to ensure that the depth at corner pixels is as accurate as possible.   

\subsection{Optimization Algorithm} 
In this section, we describe, step by step, how our calibration scheme works. Algorithm \ref{our_calib_algorithm} delineates our depth based calibration process.

\begin{algorithm}[t!]
 \begin{algorithmic}[1]
\Procedure{DepthBasedCalib}{$\bm{x}_w$, $\bm{x}_p$, $\bm{d}$, $cSize$}   
 \State{$\bm{V} \gets colorCalib(\bm{x}_w,\bm{x}_p)$}
 \State $\hat{\bm{d}}\gets planarizeDepth(\bm{x}_p,\bm{d},\bm{K},\bm{k}_c)$
 \State $count \gets 0$
 \While{($count\leq maxIter ~~\& ~~ \epsilon \geq threshold$)}
 \State $\hat{\bm{K}}\gets updateK(\bm{x}_p,\hat{\bm{d}},\bm{K},\bm{k}_c)$
 \State $\hat{\bm{d}}\gets planarizeDepth(\bm{x}_p,\hat{\bm{d}},\hat{\bm{K}},\bm{k}_c)$
 \State $\epsilon \gets errorIn3D(\bm{x}_p,\hat{\bm{d}},\hat{\bm{K}},\bm{k}_c,cSize)$
 \State $count\gets count+1$
 \EndWhile
 \State $\hat{\bm{R}}^j, \hat{\bm{t}}^j \gets localOptim(\bm{x}_w,\bm{x}_p,\hat{\bm{d}},\hat{\bm{K}},\bm{k_c},\bm{R}^j, \bm{t}^j)$
 \State $\bm{\hat{k}_c}\gets updateDistortion(\bm{x}_w,\bm{x}_p,\hat{\bm{d}},\hat{\bm{K}},\bm{k}_c)$
  \State $\bm{\hat{V}}\gets globalOptim(\bm{x}_w,\bm{x}_p,\hat{\bm{d}},\bm{V})$
 \State \textbf{return} $\bm{\hat{V}}$ 
 \EndProcedure
 \end{algorithmic}
 \caption{Depth based calibration}
 \label{our_calib_algorithm}
\end{algorithm}

\vspace{1mm}
\noindent \textbf{Color Image Calibration (line 2):} We perform traditional calibration as described in Section \ref{Sec_std_calib}. This provides us an initial estimate for the calibration parameters.

\begin{figure}[t!]
\centering
\subfigure[]{
\includegraphics[scale=0.31]{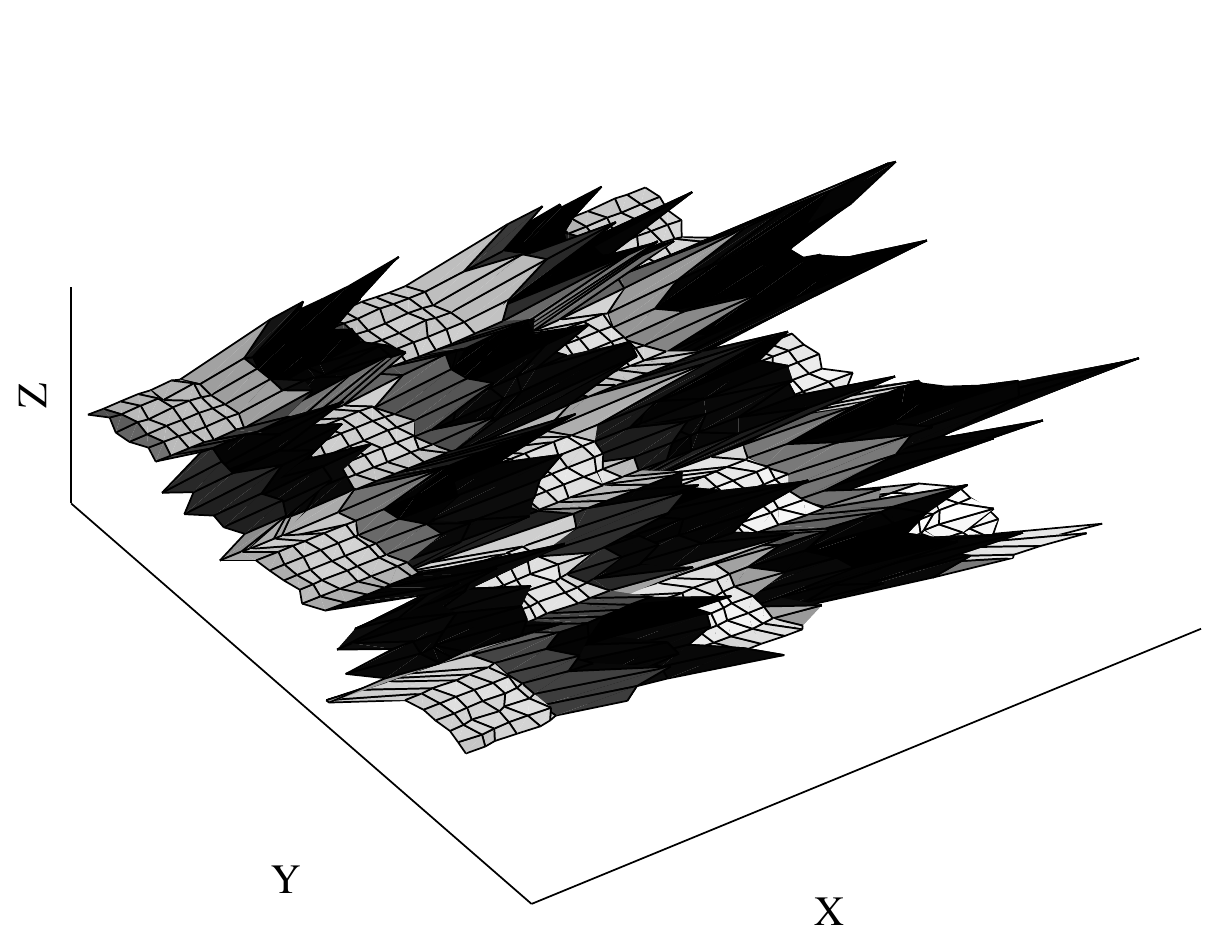}
\label{Fig:checkerboard_3d_ori} 
}
\subfigure[]{
\includegraphics[scale=0.31]{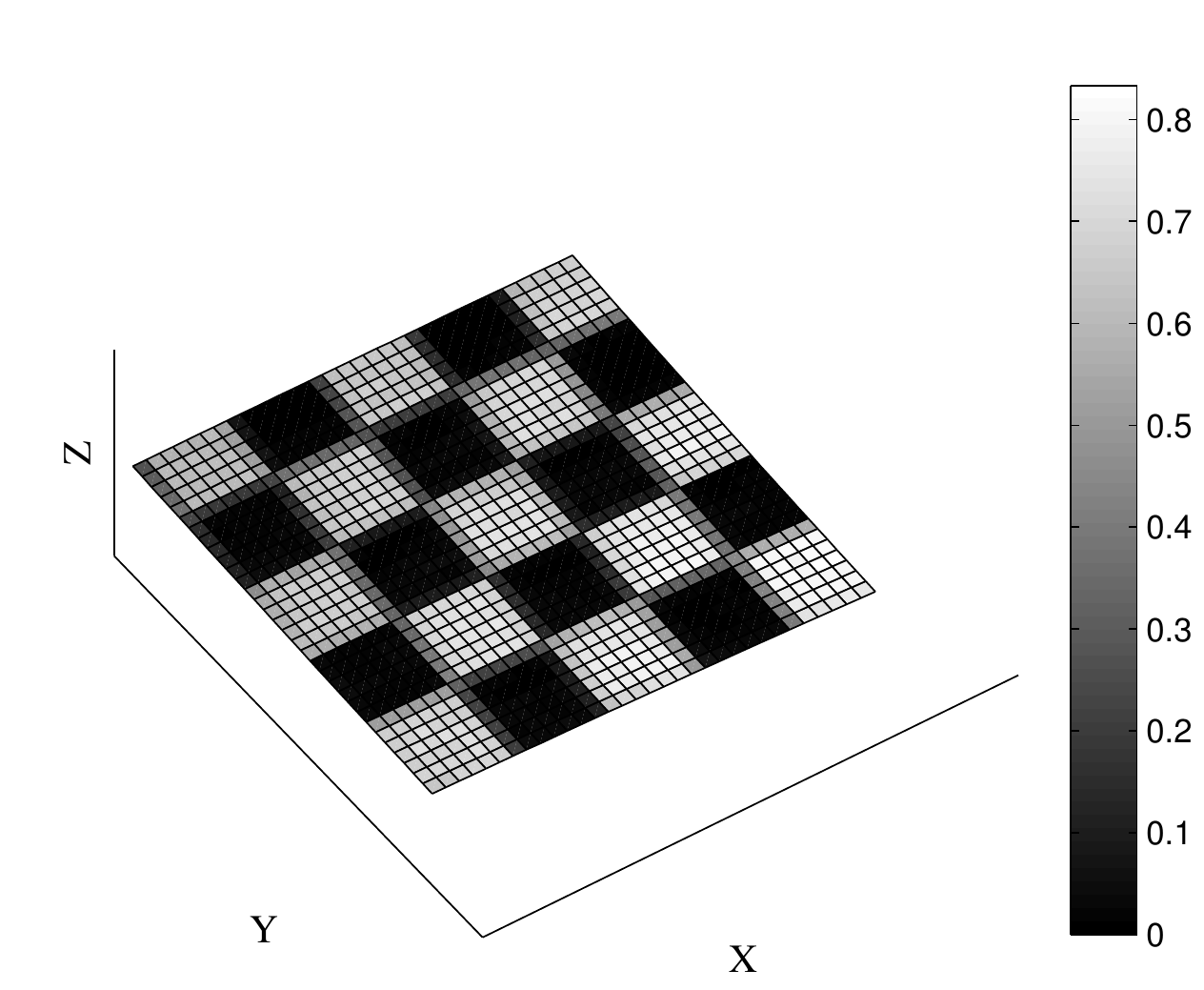}
\label{Fig:checkerboard_3d_planarized} 
}
\caption{\label{checkerboards_3D} Checkerboards projected in 3D using a) Original depth information  b) 3D planarization. }
\end{figure}

\vspace{1mm}
\noindent \textbf{Planarizing the depth image (line 3):} Since we only look at the interior corner points of a planar checkerboard, there is insignificant flying-pixel noise. Instead of denoising the depth measurement through spatial filtering, we employ prior knowledge about the scene which is a checkerboard in our case. We account for wiggling error and reflectivity based noise by performing image segmentation and 3D plane estimation. We use the corner pixel information to segment out the white squares where depth is more accurate than the black squares. This is because the Poisson-shot noise is higher in darker regions (black squares) compared to lighter regions (white squares) as seen in Fig. \ref{Fig:checkerboard_3d_ori}. We segment out the white squares and use their corresponding depth along with initial calibration parameter estimates to project the points in 3D. Thereafter, we use RANSAC along with gradient threshold to find the best plane using SVD. We estimate the depth at sub-pixel corners by finding the intersection of this estimated plane and a line passing through the sub pixel corners when projected in 3D using traditional calibration results. This provides us a more accurate depth at the sub-pixel corners as seen in Fig. \ref{Fig:checkerboard_3d_planarized}. The wiggling error is non-systematic and can lead to both under and over estimation of depth \cite{lindner2010time}. We claim that the 3D planarization eliminates the wiggling error in these regions once we have enough white checkerboard regions. We denote this denoised depth as $\hat{\bm{d}}$.

\begin{figure*}[t!]
\centering
\includegraphics[scale=0.52]{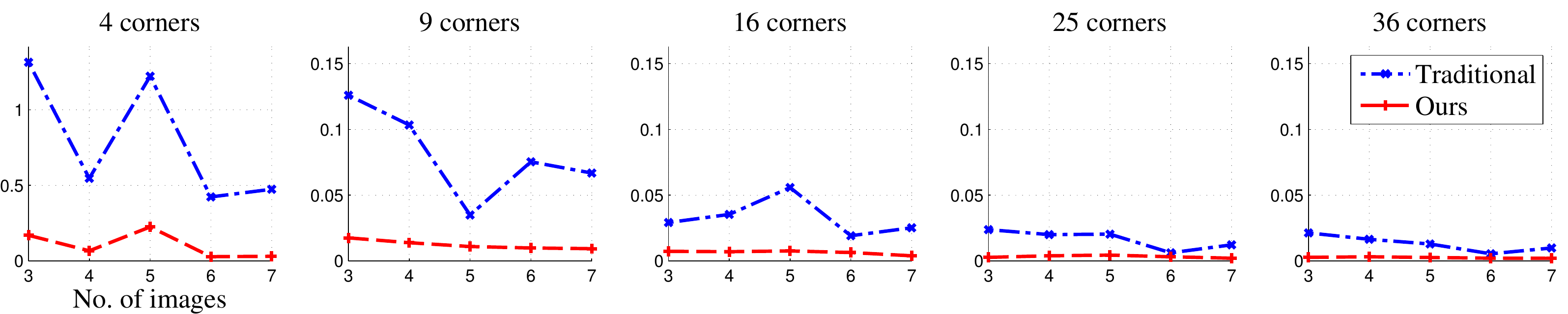}
\caption{Relative error in focal length for noisy synthetic data. We magnified the scale of the vertical axis in the cases of $9$-$36$ corners to highlight the accuracy of our calibration scheme.}
\label{Fig:synthetic_data_calib_results}
\end{figure*}

\begin{figure*}[t!]
\centering
\includegraphics[scale=0.52]{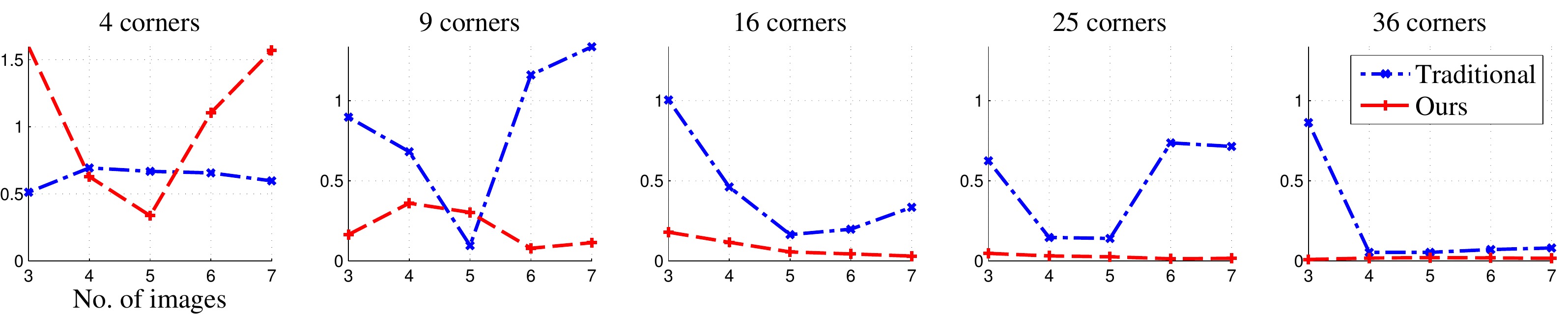}
\caption{Relative error in focal length for PMD depth camera.}
\label{Fig:PMD_data_calib_results}
\end{figure*}

\vspace{1mm}
\noindent \textbf{Updating $\bm{K}$ (lines 6-8):} The calibration parameters provided by traditional calibration when using a small set of images and corners are very unreliable. Since calibration procedure involves using non-linear estimation, a good initialization of the calibration parameters is extremely important. Hence, it is critical to re-initialize these parameters before using them for global optimization. Due to the coupling of $\bm{K}$ with $\bm{R}^j$ and $\bm{t}^j$, as seen in Eqs. (\ref{2d_projection}-\ref{eq_world_coord}), traditional calibration often fails to provide a good estimate for intrinsic calibration matrix as we lose a degree of freedom by projecting 3D coordinates onto the 2D camera plane. First, we use the estimated distortion parameters to obtain the normalized pixel positions for each corner, $\bm{x}_n$. We use the denoised depth, $\hat{\bm{d}}$, to obtain the 3D coordinates for each corner by projecting 2D corner locations in 3D: 
\begin{equation}
\bm{x}_c = \frac{\hat{\bm{d}}}{\|h^{-1}(\bm{K}^{-1}\bm{x}_p)\|_2}h^{-1}(\bm{K}^{-1}\bm{x}_p)
\end{equation}
We use a non linear optimizer to re-estimate $\bm{K}$ by enforcing projected checkerboard squares in 3D using the denoised depth data to be the same size as the actual checkerboard squares for each image:
\begin{equation}
\hat{\bm{K}} = \operatornamewithlimits{argmin}\limits_{\bm{K} } \sum_{i=1}^M\sum_{l \in  N(i)} {\left( \|\bm{x}_c^i - \bm{x}^l_c\|_2-cSize \right)^2}
\end{equation}
where $cSize$ refers to the checkerboard square size and $N(i)$ represents the neighbors of $i^{th}$ corner. We repeat this process red until at least $50\%$ of the images have an avg. $3D$ distance between points to be within $20\%$ of the checkerboard size. This provides us a reliable initial estimate for $\bm{K}$ which is crucial for the optimization process.

\vspace{1mm}
\noindent \textbf{Re-initialization (lines 11-12): } We use the updated $\bm{K}$ to re-initialize our extrinsic parameters in the same fashion as it is done for traditional calibration process. We also update the distortion parameters by assuming the remaining parameters as groundtruth and minimizing the objective function in Eq. (\ref{our_opti_eq}).

\vspace{1mm}
\noindent \textbf{Global Optimization (line 13): } Finally, we bundle everything together and perform a global optimization using Eq. (\ref{our_opti_eq}) using LMA as our non-linear solver with our new Jacobian matrix.

\section{Experimental results}
In this section, we perform synthetic and real experiments on PMD camera and compare our calibration scheme with the traditional calibration scheme.

\vspace{1mm}
\noindent \textbf{Synthetic data results: } We synthesized a $12\times12$ checkerboard with $50mm$ checker size containing $121$ interior corners. We used upto $7$ images and $36$ corners for calibration. We added white Gaussian noise to corner pixels and depth data with a standard deviation of $0.01$ pixels and $10mm$ respectively to generate noisy data. This amount of noise resembles the noise present in real data in corner estimation and depth measurements captured by PMD cameras. We used varying subsets of $7$ images and $36$ corners to estimate the calibration parameters to highlight the fact that our approach outperforms traditional approach when little information is available for calibration. We tested the calibration results on the entire checkerboard region ($121$ corners). Both the traditional and our calibration approaches achieved perfect results for noiseless dataset when more than $9$ corners and $3$ images are available. Table \ref{table_noisy} shows the mean 3D error as shown in Eq. (\ref{3d_error}) between the groundtruth corners and corners computed using the estimated calibration parameters from the two methods and groundtruth depth. Our approach outperforms traditional calibration in every test. Fig. \ref{Fig:synthetic_data_calib_results} shows relative error in focal length (=$\frac{|\Delta\bm{f}|}{\bm{f}_g}$) for noisy synthetic data. Our approach consistently provided significantly better results than the traditional calibration approach. We observed similar improvements in optical center and extrinsic calibration parameters. 
\begin{equation}\label{3d_error}
\epsilon =\frac{1}{MN}\sum_{j=1}^N\sum_{i=1}^M \|\hat{\bm{x}}^{i,j}_w - {\bm{x}}^{i,j}_w  \|_2 
\end{equation}

\noindent \textbf{Real data (PMD) results: } We used a checkerboard with $50mm$ checker size to capture $12$ images using a PMD camera. Each checkerboard contains $70$ corners. We used upto $7$ images and $36$ corners to estimate the intrinsic and extrinsic calibration parameters. We compare the focal length, $\bm{f}$, obtained from both approaches to the manufactured focal length of the PMD camera, $[284.4,284.4]^\top$ pixels. We assume this value as groundtruth.  As seen in Fig. \ref{Fig:PMD_data_calib_results}, our approach consistently provides a reasonably accurate focal length while  traditional calibration estimates a highly inaccurate focal length in most cases. One significant deviation from this behaviour happens when only four corners are available for calibration. This is because the estimation process  diverges as the initial estimates are far away from the ground truth where the non-linear estimation process (LMA) is known to fail frequently. However, once we use nine or more corners per image, our approach provides significantly better results consistently.

\section{Conclusion}
We presented a simple and accurate method to simultaneously denoise depth data and calibrate depth cameras. The presented method excels in estimating calibration parameters when only a handful of corners and calibration images are available where traditional approach really struggles. While this approach is simple and easily applicable, it still relies on using a checkerboard pattern to perform calibration. In future, we intend to exploit planarity in generic scene to perform calibration so that any user at home can use our calibration procedure. 


\begin{table}[t!]\footnotesize
\centering
\resizebox{8.5cm}{!} {
\begin{tabular}{| c | c || c | c | c | c | c | } \hline
 \multicolumn{2}{|c||}{\diagbox{$\#$ corners}{$\#$ images} } & 3 & 4 & 5 & 6 & 7\\  \hline  \hline 
  \multicolumn{1}{ |c| }{\multirow{2}{*}{4}} & traditional & 53.6795  & 12.5422   & 8.6423  &  3.1773  &  1.8796 \\ \cline{2-7} 
  \multicolumn{1}{ |c| }{}  & ours & 7.9059 &   1.7206  &  0.8978  &  0.6689   & 0.4144 \\  \hline 
  \multicolumn{1}{ |c| }{\multirow{2}{*}{9}}  & traditional &  51.8946 &  10.7673 &   2.7336  &  3.3252  &  2.1771  \\ \cline{2-7} 
  \multicolumn{1}{ |c| }{} & ours & 8.0193  &  1.7345  &  0.8591  &  0.4818   & 0.4220 \\  \hline 
 \multicolumn{1}{ |c| }{\multirow{2}{*}{16}}  & traditional &  151.9972 &   8.2549   & 6.6561  &  3.2660  &  1.7817 \\  \cline{2-7}
  \multicolumn{1}{ |c| }{} & ours & 37.3031  &  1.7453 &   0.8137 &   0.5875   & 0.4384  \\ \hline
\multicolumn{1}{ |c| }{\multirow{2}{*}{25}}  & traditional &  57.8784  &  8.7056   & 5.2105  &  2.5943  &  0.6113  \\ \cline{2-7} 
  \multicolumn{1}{ |c| }{} & ours &  4.0092  &  1.5235  &  0.8008  &  0.5373   & 0.3934 \\  \hline 
 \multicolumn{1}{ |c| }{\multirow{2}{*}{36}}  & traditional & 40.4462  &  8.1659   & 2.9612 &   1.3889  &  0.8504  \\ \cline{2-7} 
  \multicolumn{1}{ |c| }{} & ours  & 4.2415 &   1.7081  &  0.8109  &  0.5640   & 0.4449 \\  \hline 
\end{tabular}}
\caption{Avg. 3D error between groundtruth corners and projected corners in $mm$}
\label{table_noisy}
\end{table}

\vfill
\pagebreak
\label{sec:ref}

\bibliographystyle{IEEEbib}
\bibliography{biblio}

\begin{thebibliography}{10}

\bibitem{kinect_website}
Microsoft,
\newblock ``{K}inect for {X}box,'' 2013.

\bibitem{pmd_website}
{PMDT}echnologies,
\newblock ``{PMD} camcube 3.0,'' 2013.

\bibitem{wavi_xtion_website}
ASUS,
\newblock ``Wavi xtion: Intuitive living room experience,'' 2013.

\bibitem{Kolb2009}
Andreas Kolb, Erhardt Barth, Reinhard Koch, and Rasmus Larsen,
\newblock ``Time-of-flight sensors in computer graphics,''
\newblock in {\em Proc. Eurographics (State-of-the-Art Report)}, pp.
  119--134, 2009.

\bibitem{Izadi2011}
Shahram Izadi, Richard~A. Newcombe, Kim, et~al.,
\newblock ``Kinect{F}usion: real-time dynamic {3D} surface reconstruction and
  interaction,''
\newblock  ACM. SIGGRAPH, pp. 23:1--23:1, 2011.

\bibitem{shotton2013real}
Jamie Shotton, Toby Sharp, Kipman, et~al.,
\newblock ``Real-time human pose recognition in parts from single depth
  images,''
\newblock {\em Communications of the ACM}, vol. 56, no. 1, pp. 116--124, 2013.

\bibitem{zhang2000flexible}
Zhengyou Zhang,
\newblock ``A flexible new technique for camera calibration,''
\newblock {\em IEEE Transactions on Pattern Analysis and Machine Intelligence
  (TPAMI)}, vol. 22, no. 11, pp. 1330--1334, 2000.

\bibitem{Maybank1992}
Stephen~J. Maybank and Olivier~D. Faugeras,
\newblock ``A theory of self-calibration of a moving camera,''
\newblock {\em International Journal of Computer Vision}, vol. 8, pp. 123--151,
  1992.

\bibitem{kahlmann2006calibration}
Timo Kahlmann, Fabio Remondino, and H~Ingensand,
\newblock ``Calibration for increased accuracy of the range imaging camera
  {S}wissranger$^{TM}$,''
\newblock {\em Image Engineering and Vision Metrology (IEVM)}, vol. 36, no. 3,
  pp. 136--141, 2006.

\bibitem{lindner2006lateral}
Marvin Lindner and Andreas Kolb,
\newblock ``Lateral and depth calibration of pmd-distance sensors,''
\newblock in {\em Advances in Visual Computing}, pp. 524--533. Springer, 2006.

\bibitem{fuchs2008extrinsic}
Stefan Fuchs and Gerd Hirzinger,
\newblock ``Extrinsic and depth calibration of tof-cameras,''
\newblock in {\em Computer Vision and Pattern Recognition (CVPR)}. IEEE, 2008.

\bibitem{beder2008calibration}
Christian Beder and Reinhard Koch,
\newblock ``Calibration of focal length and 3d pose based on the reflectance
  and depth image of a planar object,''
\newblock {\em International Journal of Intelligent Systems Technologies and
  Applications}, vol. 5, no. 3, pp. 285--294, 2008.

\bibitem{kim2011depth}
Sung-Yeol Kim, Woon Cho, Andreas Koschan, and Mongi~A Abidi,
\newblock ``Depth data calibration and enhancement of time-of-flight
  video-plus-depth camera,''
\newblock in {\em Future of Instrumentation International Workshop (FIIW)}.
  IEEE, pp. 126--129, 2011.

\bibitem{herrera2011accurate}
Daniel Herrera, Juho Kannala, and Janne Heikkil{\"a},
\newblock ``Accurate and practical calibration of a depth and color camera
  pair,''
\newblock in {\em Computer Analysis of Images and Patterns}. Springer, 
  pp. 437--445, 2011.

\bibitem{herrera2012joint}
Daniel Herrera, Juho Kannala, Janne Heikkil{\"a}, et~al.,
\newblock ``Joint depth and color camera calibration with distortion
  correction,''
\newblock {\em IEEE Transactions on Pattern Analysis and Machine Intelligence
  (TPAMI)}, vol. 34, no. 10, pp. 2058--2064, 2012.

\bibitem{bouguet2004camera}
Jean-Yves Bouguet,
\newblock ``Camera calibration toolbox for matlab,''
\newblock 2004.

\bibitem{Hartley2004}
Richard Hartley and Andrew Zisserman,
\newblock {\em Multiple {V}iew {G}eometry in computer vision}, vol.~2,
\newblock Cambridge Univ Press, 2000.

\bibitem{Bradski2008}
Gary Bradski and Adrian Kaehler,
\newblock {\em Learning OpenCV: Computer vision with the OpenCV library},
\newblock O'reilly, 2008.

\bibitem{lindner2010time}
Marvin Lindner, Ingo Schiller, Andreas Kolb, and Reinhard Koch,
\newblock ``Time-of-flight sensor calibration for accurate range sensing,''
\newblock {\em Computer Vision and Image Understanding}, vol. 114, no. 12, pp.
  1318--1328, 2010.

\end{thebibliography}

\end{document}